\documentclass[journal]{IEEEtran}

\usepackage{xcolor,soul,framed} %,caption

\colorlet{shadecolor}{yellow}
\usepackage[pdftex]{graphicx}
\graphicspath{{../pdf/}{../jpeg/}}
\DeclareGraphicsExtensions{.pdf,.jpeg,.png}

\usepackage[cmex10]{amsmath}
\usepackage{array}
\usepackage{mdwmath}
\usepackage{mdwtab}
\usepackage{eqparbox}
\usepackage{url}
\usepackage{makecell}
\usepackage{booktabs}
\usepackage{multirow}
\usepackage{amssymb}

\usepackage{etoolbox}
\makeatletter
\patchcmd{\@makecaption}
  {\scshape}
  {}
  {}
  {}
\makeatother

\begin{document}
\bstctlcite{IEEEexample:BSTcontrol}
    \title{Learning based Deep Disentangling Light Field Reconstruction and Disparity Estimation Application}
  \author{Langqing Shi, Ping Zhou
      }% <-this % stops a space

\markboth{IEEE TRANSACTIONS ON MICROWAVE THEORY AND TECHNIQUES, VOL.~60, NO.~12, DECEMBER~2012
}{Roberg \MakeLowercase{\textit{et al.}}: High-Efficiency Diode and Transistor Rectifiers}

% ====================================================================
\maketitle

% === ABSTRACT ====================================================================
% =================================================================================
\begin{abstract}
%\boldmath
Light field cameras have a wide range of uses due to their ability to simultaneously record light intensity and direction. The angular resolution of light fields is important for downstream tasks such as depth estimation, yet is often difficult to improve due to hardware limitations. Conventional methods tend to perform poorly against the challenge of large disparity in sparse light fields, while general CNNs have difficulty extracting spatial and angular features coupled together in 4D light fields. The light field disentangling mechanism transforms the 4D light field into 2D image format, which is more favorable for CNN for feature extraction. In this paper, we propose a Deep Disentangling Mechanism, which inherits the principle of the light field disentangling mechanism and further develops the design of the feature extractor and adds advanced network structure. We design a light-field reconstruction network (i.e., DDASR) on the basis of the Deep Disentangling Mechanism, and achieve SOTA performance in the experiments. In addition, we design a Block Traversal Angular Super-Resolution Strategy for the practical application of depth estimation enhancement where the input views is often higher than 2x2 in the experiments resulting in a high memory usage, which can reduce the memory usage while having a better reconstruction performance. 
\end{abstract}

% === KEYWORDS ====================================================================
% =================================================================================

% \fi
\IEEEpeerreviewmaketitle

% ====================================================================
% ====================================================================
% ====================================================================

% === I. INTRODUCTION =============================================================
% =================================================================================
\section{Introduction}

Light field cameras have a wide range of applications due to their ability to record both the intensity and direction of light. A light field camera encodes a 3D scene into a 4D light field and reconstructs it, where there are 2 dimensions for the space and 2 dimensions for the angle. This property allows the light field to enable many valuable applications, such as depth estimation \cite{sepi, wang2016depth}, post-shot refocusing \cite{refocusing1, refocusing2}, 3D reconstruction \cite{zhou2022phase}, and so on. Light field cameras can be categorized according to the principle as micro-lens array based, gantry based \cite{stf_gantry}, camera array based \cite{camera_array}, and so on. Among the light field camera solutions based on various principles, the one based on micro-lens arrays is the least expensive and the most suitable for practical applications. However, although micro-lens arrays separate light rays from different sub-apertures, these light rays still share the same CCD, which means that it is difficult to retain both spatial and angular resolution in a light-field camera. Specifically, the user must trade-off between recording sharper images with a smaller number of views and blurrier images with a larger number of views. To solve this problem, spatial super-resolution algorithms and angular super-resolution algorithms have been proposed. Among them, angular super-resolution algorithms are important for light field downstream tasks such as depth estimation, and this paper addresses light field angular super-resolution algorithms.

Some traditional algorithms try to perform new view synthesis \cite{shi_fourier, wanner2013variational}, however, these algorithms suffer from sparsity and often perform poorly in scenes with large disparity. Nowadays, with the continuous development of deep learning technology, many learning-based algorithms for light field angle super-resolution have been proposed. The commonality of these algorithms is that the dense light field ground truth is used as the supervision, and the error minimization of the dense light field after super-resolution is used as the optimization objective to train the super-resolution model. Some methods \cite{kalantari2016learning, lfasr, lfasr_fs_gaf} first estimate the disparity of each view by a prefixed depth estimation module, and then warp and refine the input views to the to-be-demanded views based on the disparity map by a warping module jointly trained with the depth estimation module. Some other methods \cite{sheared_epi, lfepicnn} reconstruct the light field parallax structure by extracting the linear slope features in the epipolar image based on the correlation between the linear slope in the epipolar image and the depth of the spatial object points. Wang et al. proposed a light-field decoupling mechanism \cite{distg_asr} to flatten the 2 angular and 2 spatial dimensions of the light field into a 2D planar image in the form of a macro-pixel image, which facilitates the learning of the light-field parallax structure by a convolutional network.

As a bridge between upstream tasks such as light field calibration and downstream tasks such as depth estimation, light field angular super-resolution is closely related to practical applications. In practical applications, the number of input views is not as low as $2 \times 2$, which is commonly used in most experiments of light field angle super-resolution methods, but more commonly $5 \times 5$ or even higher. The memory usage of the learning-based light-field angular super-resolution method is positively and approximately linearly related to the number of input views. Such a property leads to a significant increase in the model's memory usage as the input views become relatively dense. We tested a variety of learning-based methods, and the memory usage at $5 \times 5$ input views is close to 6 times that of $2 \times 2$, which places high demands on the hardware.
The main innovation of this paper is to propose a sparse light field reconstruction method based on a light field disentangling mechanism, which we refer to as the Deep Disentangling Mechanism. The deep disentangling mechanism inherits the principle of the light field disentangling mechanism \cite{distg_asr} and makes several improvements to address the challenge of large disparity in sparse light fields, which include adjusting the underlying structure of the feature extractor and incorporating advanced network structures. We designed the DDASR network and extended applications based on the Deep Disentangling Mechanism. 
In summary, the contributions of this paper are as follows:
\begin{itemize}
    \item We propose a Deep Disentangling Mechanism, which can better learn the light field parallax structure than the original light field disentangling mechanism.
    \item We design a DDASR network based on the Deep Disentangling Mechanism, train the network on a variety of light field datasets (both synthetic and real scenes), and compare it with other SOT Amethods. The experimental results show that its performance reaches the state-of-the-art.
    \item We propose the Block Traversal ASR Strategy (BTAS) on the basis of DDASR, which effectively solves the problem of excessive memory usage in practical applications with middle views number, and achieves better performance than direct angular super-resolution under the condition of limiting the memory size.
\end{itemize}
The remainder of this paper is organized as follows. In Section II we summarize the various approaches in the field of learning-based light field angular super-resolution and their advantages and disadvantages. In Section III we present the principle of the Deep Disentangling Mechanism and the design details of the DDASR network. The implementation details and results of the ablation and comparison experiments will be illustrated in Section IV, and the depth estimation enhancement extended application based on the Block Traversal ASR Strategy will be detailed in Section V. Finally, we conclude in Section VI.

% === II. Harmonically-Terminated Power Rectifier Analysis ========================
% =================================================================================
\section{Related Work}

Many methods have been proposed for the problem of light field angular super-resolution. Among the non-learning methods, Shi \cite{shi_fourier} et al. performed light field reconstruction using the sparsity representation of the light field after Fourier transform. Wanner et al. in \cite{wanner2013variational} regarded the acquisition of the disparity map of the scene as an energy-minimization problem with a full-variance a priori. The disparity map is obtained by global optimization of the structure tensor in the epipolar image, and finally the disparity map is used as the basis for new viewpoint synthesis. However, these non-learning methods perform poorly in the face of complex scenes (e.g., non-Lambertian surfaces and occluded regions) as well as sparse light fields with large disparity. In these challenging scenarios, learning-based methods rely on the strong expressive ability of the deep learning model to achieve better performance. Learning-based light field angular super-resolution is mainly categorized into explicit depth-dependent and depth-independent.

\subsection {Light field Angular Super-Resolution Methods with Explicit Depth Estimation}

This class of methods roughly divides the whole model into two phases, with the first phase performing scene depth estimation from light field inputs, and the second phase performing view synthesis based on the scene depth obtained in the first phase. Wu et al \cite{sheared_epi} proposed a method based on the correspondence between the slope of a line in the epipolar image and the depth of an object point in space. They moved each row of pixels in the epipolar image to determine the disparity of each point in a given search space, which was used as the basis for view synthesis. Kalantari \cite{kalantari2016learning} decomposed the light field angular super-resolution task into two phases, depth estimation and color estimation, and used an end-to-end network to integrate the two phases. Jin et al. proposed a coarse-to-fine (coarse-to-fine) network framework in \cite{lfasr}, where the scene depth is first obtained by the depth estimation module, then the new view is obtained by warping the input views according to the scene depth. Finally, the new view is refined by the refinement module. Jin et al. further developed this framework in the subsequent \cite{lfasr_fs_gaf} by updating their network design and using a PSV-based input form, which allows the network to accept input views from arbitrary pattern instead of fixed positions. Shi \cite{shi2020learning} et al. noticed that the pixel domain is helpful for reconstruction of low frequency features of an image while the feature domain is helpful for reconstruction of high frequency features. They designed two reconstruction modules in the pixel domain and the feature domain based on the above rule, and their depth estimation is realized by a lightweight optical flow estimation module.

Such methods with explicit depth estimation have a common problem that the accuracy of depth estimation seriously affects the quality of the final viewpoint synthesis. At the same time, from the perspective of engineering practice, the light field data with full viewpoints in the second stage occupies a tremendous amount of memory, which limits the network depth.

\subsection {Light field Angular Super-Resolution Methods without Explicit Depth Estimation}

This type of approach, compared to the one in Section 2.1, does not have an explicit module for depth estimation in the model, but instead learns the light field parallax structure through deeper network layers, and finally obtains the new viewpoints through some kind of up-sampling mechanism (e.g., convolution and pixel shuffle). Yoon et al. proposed LFCNN \cite{yoon2015learning} and implemented both light field angular super-resolution and spatial super-resolution on it, by first doing spatial super-resolution for each SAI and then synthesizing new views based on its surrounding views. Wu et al. proposed that the synthesis of the new perspective is equivalent to reconstructing the missing rows between each row of the polar plot, and proposed a blurring-reconstruction-deblurring framework for light field reconstruction with CNN as the core after a frequency domain analysis of the reconstruction process \cite{lfepicnn}. Yeung et al. argued that the traditional convolution on a single 2D SAI cannot extract the parallax structure specific to the light field, but the cost of 4D convolution on a 4D light field is too huge, so they proposed to replace the 4D convolution with the equivalent of alternating 2D convolutions of different dimensions, called pseudo-4D convolution \cite{p4dcnn}. Yang et al. \cite{yang2023light} explicitly extracted structural and scene information of the light field through an end-to-end network, respectively. Wang et al. proposed a MacPI-based light field decoupling mechanism \cite{distg_asr}, which flattens a 4D light field into a 2D macro-pixel image. They realized light field angular super-resolution, spatial super-resolution, and depth estimation on this basis, respectively.

The method without explicit depth estimation does not suffer from depth estimation error and is capable of constructing large-scale network models.

\section{Method}

In this section, we apply the deep disentangling mechanism introduced in the previous section to our proposed network and describe its structure in detail. The experimental results of the network will be shown in Section 4. 

\subsection{The Deep Disentangling Mechanism}
\subsubsection{Light Field Disentangling Mechanism}

\begin{figure}
  \begin{center}
  \includegraphics[width=3.0in]{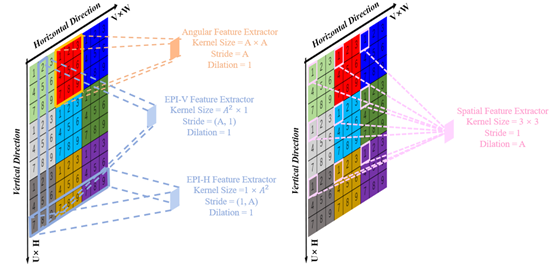}\\
 \caption{Schematic diagram of spatial feature extractor, angular feature extractor, and EPI feature extractor. In the figure, the angular dimension U = V = A = 3 and the spatial dimension H = W = 3. Pixels of the same color in the macro-pixel image in the legend come from the same spatial position, while pixels with the same number come from the same angular position. The left figure contains the angular feature extractor and the two EPI feature extractor for both vertical and horizontal, while the right figure shows the spatial feature extractor}\label{fig:feature_extractor}
  \end{center}
\end{figure}

Wang et al. proposed a complete light field disentangling mechanism in \cite{distg_asr} and applied it to light field super-resolution (both special and angular), and depth estimation, all of which achieved leading performance. The light field disentangling mechanism first transforms the 4D light field into a macro-pixel image (MacPI), i.e., pixels with the same spatial position are extracted from all viewpoints to form a macro-pixel, and these macro-pixels are stitched together into a macro-pixel image according to their corresponding spatial positions. Subsequently, Wang et al. designed three types of feature extractors, namely spatial feature extractor, angular feature extractor and EPI feature extractor, as shown in Figure \ref{fig:feature_extractor}

\textbf{Spatial Feature Extractor}: The SFE extracts spatial information by applying a convolution kernel with size=$3 \times 3$, stride=1, dilation=A to the MacPI. The spatial feature extractor acts on pixels from the same viewpoint within different macro-pixels (i.e., within a spatial neighborhood), which do not contain angular information between them, but only spatial information.

\textbf{Angular Feature Extractor}: The AFE extracts angular information by applying a convolution kernel with size=$3 \times 3$, stride=A, dilation=1 to the MacPI. The angular feature extractor acts on pixels from neighboring viewpoints within the same macro-pixel (at the same spatial location), which contain no spatial information between them, only angular information.

\textbf{EPI Feature Extractor}: We designed epi feature extractor for vertical and horizontal directions respectively. In the horizontal direction, for example, the correlation between angular and spatial information, which is the information characterized by the epi, is extracted by applying a convolution kernel with size=$1\times A^2$, dilation=1, and step sizes of 1 and $A^2$ in the vertical and horizontal directions, respectively. This epi feature extractor acts on a number of pixels in the horizontal direction within a number of macro pixels in the horizontal direction, which is equivalent to the epi obtained by fixing both the angular and spatial coordinates in the horizontal direction in the 4D light field, and these pixels contain the correlation between the angular features and the spatial features. The same is true for the vertical one. 

\subsubsection{The Deep Disentangling Mechanism}
Inspired by Wang et al \cite{distg_asr}, we apply the light field disentangling mechanism to light field angular super-resolution and make several improvements based on it. We refer to the improved light field disentangling mechanism and its accompanying network structure design optimization as the deep disentangling mechanism. These improvements include:
\begin{itemize}
    \item Replace the spatial and angular feature extractors with a spatial feature extraction block and an angular feature extraction block, respectively. These two feature extraction blocks are combined from the corresponding feature extractors, and a layer-by-layer concatenate structure is added to increase the depth of the network.
    \item Incorporating channel attention with the layer-by-layer concatenate structure to improve the learning ability of the network.
    \item  Adjusting the ratio of each stage of the network.
    \item Add $3 \times 3$ convolution after the angular feature extractor to cope with the challenge of large parallax caused by sparse light field.
\end{itemize}

\begin{figure}
  \begin{center}
  \includegraphics[width=3.0in]{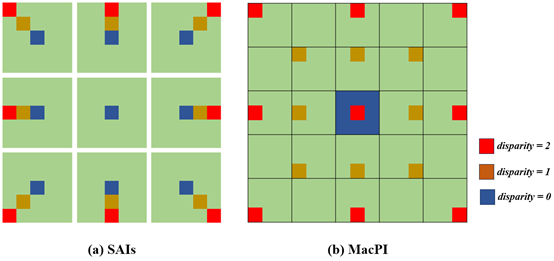}\\
 \caption{A toy example of a sparse light field. (a) is an array of subaperture images with an angular domain size of $3 \times 3$ and a spatial domain size of $5 \times 5$, and (b) is the corresponding macro-pixel image. The blue, brown, and red squares in the figure represent pixels with parallaxes of 0, 1, and 2, respectively. It should be noted that the anchor pixel of the central viewpoint in (a) and the anchor pixel of the central macro-pixel in (b) are overlapped in three colors.}\label{fig:large_disparity}
  \end{center}
\end{figure}

The 1, 2 and 3 can be summarized as the improvement of the network design, which will be elaborated in Section 3.2, while the fourth point addresses improvements to the light-field disentangling mechanism.

Light fields with more than one pixel of parallax between neighboring viewpoints are generally referred to as sparse light fields in light field reconstruction. Qualitatively, in EPI representations, the sparseness of the light field is manifested by the fact that the straight line representing a spatial object point is no longer continuous, and the line is spaced between pixels in each row, which leads to very severe aliasing artifacts with direct interpolation. In the macro-pixel image representation, the sparsity of the light field is manifested by the fact that pixels representing the same spatial location cannot be concentrated in the same macro-pixel, but are distributed in the surrounding neighborhood of macro-pixels except for the anchor pixel, and there is at most one in each macro-pixel. As shown in Figure \ref{fig:large_disparity}

Based on the definition of parallax in light field, it can be deduced that pixels representing the same object point in space satisfy the offset between different macro-pixels in a macro-pixel image:
\begin{equation}
    \begin{cases}
    \Delta h_u = d \times (u_c - u) \\
    \Delta w_v = d \times (v_c - v)
    \end{cases}
\end{equation}

where $(\Delta h_u, \Delta w_v)$ denotes the coordinate offsets of the pixels of the same object point in space in two directions between different macro-pixels in MacPI, d denotes the parallax value, and $(u, v)$ denotes the corresponding angular coordinates. As shown in Figure \ref{fig:large_disparity}(b), the blue pixels with a parallax of 0 are all concentrated in one macro-pixel in MacPI, the brown pixels with a parallax of 1 are distributed in macro-pixels with a distance of 1 around the center macro-pixel except for the anchor pixel, and the red pixels with a parallax of 2 are distributed in macro-pixels with a distance of 2 around the center macro-pixel except for the anchor pixel. From equation 1, it can be deduced that some of the pixels with parallaxes between 0 and 1 are distributed in sub-pixel locations within the macro-pixel where the anchor pixel is located, and the remainder are distributed in sub-pixel locations within the surrounding macro-pixels.

As can be seen from the previous derivation and the figure, the receptive field of the original angular feature extractor in a sparse light field is not enough to cover all the macro-pixels distributed with the corresponding pixels from the same spatial object point, and it is difficult to establish the correlation between all the macro-pixels corresponding to the same object point in a neighborhood, which is not conducive for the angular feature extractor to learn the correct angular information. To solve this problem, we add a $3 \times 3$ convolutional layer after the angle feature extractor to learn the correlation between different macro-pixels in the angle feature map, and correlate the pixels from the same object point distributed within several macro-pixels through the correlation to obtain more complete angle information.

\begin{figure}
  \begin{center}
  \includegraphics[width=3.0in]{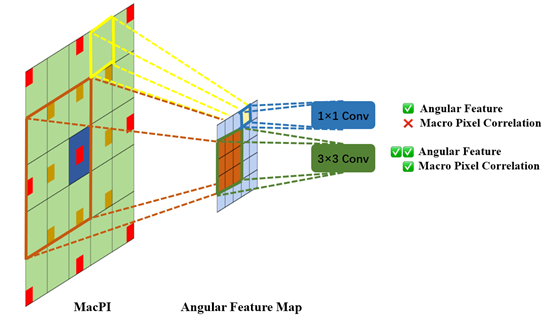}\\
 \caption{Example diagram of the improved angular feature extractor. The different colored squares in MacPI represent the same pixels as in Fig. 2. Regions boxed by different colored boxes represent different sizes of receptive fields.}\label{fig:3x3conv_afe}
  \end{center}
\end{figure}

In the example shown in Figure \ref{fig:3x3conv_afe}, the $3 \times 3$ convolutional kernel acting on the angular feature map corresponds to a larger receptive field on the MacPI, capable of covering a larger number of macro-pixels and learning correlations between pixels from the same object point contained in those macro-pixels, extracting a more complete angular feature. In contrast, the $1 \times 1$ convolution corresponds to a smaller receptive field, which can only cover a single macro-pixel and is difficult to learn the correlation between multiple macro-pixels within a neighborhood, and its extracted angle information is incomplete.

In summary, adding a $3 \times 3$ convolutional layer after the angular feature extractor is a very important improvement, although it leads to a larger number of parameters. We will try to keep the model parameter count constant by correcting the number of model modules in subsequent ablation experiments related to this improvement.

\subsection{Network Design}

\subsubsection{Overview}

The input of the network is the light field sparsely sampled in the angular domain , denoted as $LF_{sample} \in R^{UH \times VW}$. In practice, $U$ and $V$ are often equal, thus simplifying to $LF_{sample} \in R^{AH \times AW}$, where $U=V=A$. The output of the network is a dense light field super-resolved in the angular domain by a specific magnification, denoted as $\widehat{LF} \in R^{\alpha AH \times \alpha AW}$, where $\alpha$ denotes the super-resolution magnification. For example, in the common $2 \times 2 \to 7 \times 7$ task, $\alpha$ is set to $\frac{7}{2}$. The light fields described above are all represented as stitched sub-aperture images. Similar to Wang \cite{distg_asr} et al. we transform the input light field into a macro-pixel image format, i.e., $MacPI \in R^{AH \times AW}$ in order to perform spatial, angular, and EPI feature extraction. 

\begin{figure*}
  \begin{center}
  \includegraphics[width=6.0in]{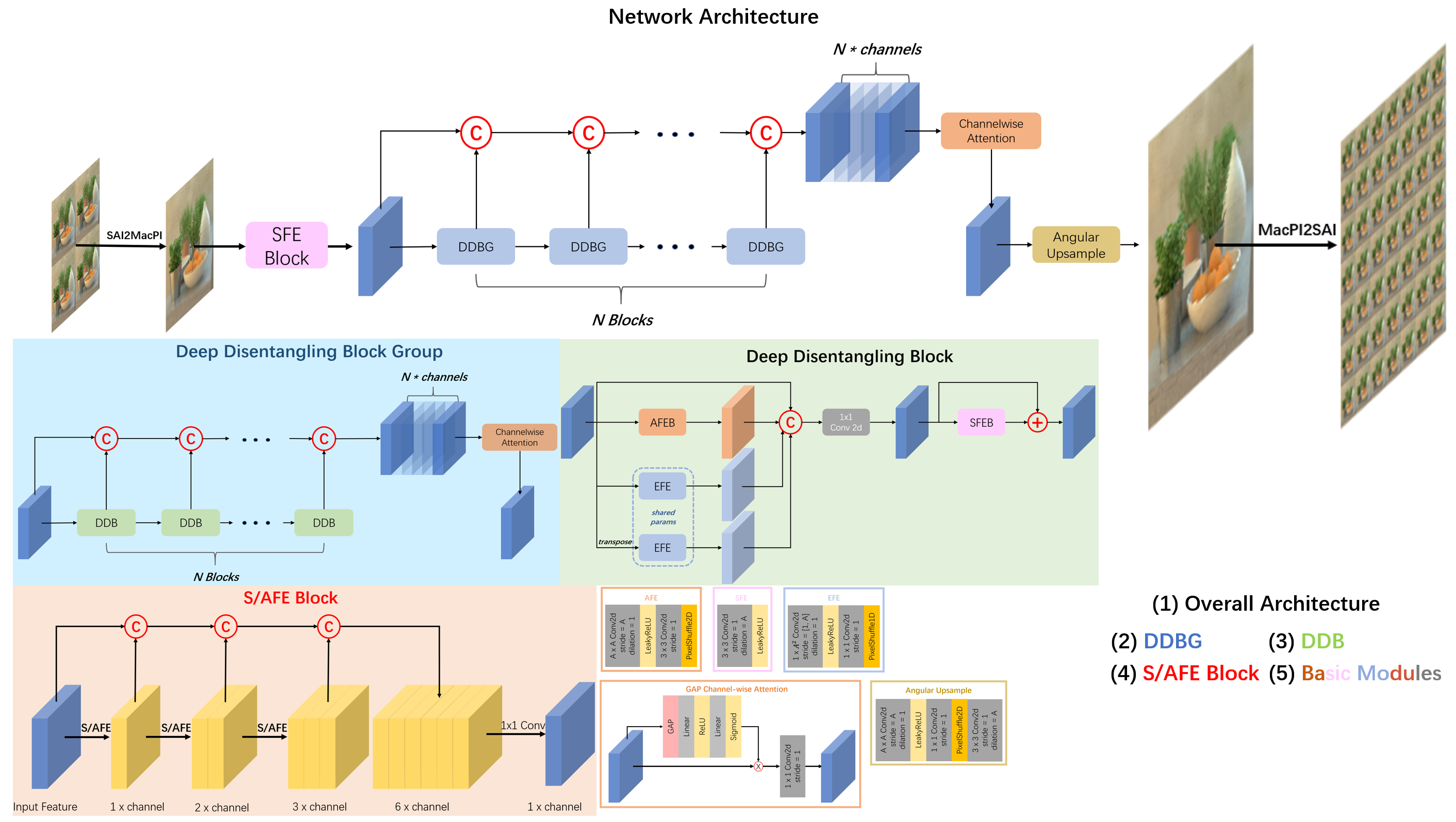}\\
 \caption{Network architect. The legend is organized as shown in the lower right corner.}\label{fig:network}
  \end{center}
\end{figure*}

The network structure is shown in Figure \ref{fig:network}. The overall architecture of the network consists of a number of Deep Disentangling Blocks Groups (hereinafter referred to as DDBG) organized by direct and skip connections, a front initial feature extraction and a final angular up-sampling module. Among them, each DDBG contains several Deep Disentangling Block (hereinafter referred to as DDB). The DDB contains a spatial feature extractor block, an angular feature extractor block, two EPI feature extractors (hereinafter referred to as SFEB, AFEB, EFE-H and EFE-V, respectively), and several other components. For easy representation and understanding, we divide the network structure into four levels:
\begin{itemize}
    \item The level 1 is the SFEB, AFEB, EFE-V, EFE-H and several other components, which form the basis of the upper level modules.
    \item The level 2 is the DDB, which efficiently combines the components from the first level in order to extract multidimensional features of the light field.
    \item The level 3 is the DDBG, which organizes DDBs in a specific way to increase the depth of the network.
    \item The level 4 portrays the top-level design of the network and contains several DDBGs to further enhance the network depth and representation capabilities. 
\end{itemize}

In the current version of the network structure, the top level contains four DDBGs, and the number of DDBs contained in each DDBG is set to 2, 2, 6, and 2, respectively. This design references the stage ratio in Swin-Transformer\cite{swin}. The ablation experiments in the next section confirm the effectiveness of this design in the light-field angle super-resolution task.

\subsubsection{Layer-by-Layer Concatenation and Channel-wise Attention}

The layer-by-layer concatenation is a design that widely used in level 4, 3 and 1. Let the input feature be $\mathcal{F}_{in} \in R^{H \times W \times C}$, and several modules connected in series are denoted as $f_1, f_2, \ldots, f_n$, respectively. These modules concatenate the outputs in the channel dimension and subsequently fuse the feature maps of the multiplicity of channels into a single one via the channel-wise attention module, the above process is denoted as $\mathcal{F}_{out}=ChnAttn[concat(f_n(f_{n-1}(\ldots(f_1(\mathcal{F}_{in})))), \ldots, f_1(\mathcal{F}_{in}))]$. The purpose of fusion is not only to perform light-field feature extraction, but also to facilitate stacking of the same structures to increase the network depth.

The channel-wise attention module first performs global average pooling on the input features $\mathcal{F}_{in} \in R^{H \times W \times C}$ to obtain the output $\mathcal{F}_{avgpool} \in R^{1 \times 1 \times C}$, followed by an MLP consisting of two linear layers and an activation function to obtain the output $\mathcal{W}_{chn} \in R^{1 \times 1 \times C}$. Finally, the output of the MLP is computed as a product of the channel dimensions of the input features to obtain the features weighted by channel-wise attention.

The ablation experiments in Section IV demonstrate the effectiveness of layer-by-layer concatenation and channel-wise attention modules. Layer-by-layer concatenation significantly reduces the number of parameters while improving the super-resolution performance by a small margin compared to the Dense connection structure common in traditional single-image super-resolution.

\subsubsection{Angular Feature Extraction Block (AFEB)}

The design of the angular feature extraction block was inspired by the work of Wang et al. [] and further improved. The angular feature extraction block contains 3 layers of angular feature convolution to extract the angle information of the light field, and the layer-by-layer channel concatenation design is applied to it. Since the angular feature extraction is located at the bottom level of the network, the extracted features are at a lower level, so the channel fusion is realized with a simple 1x1 convolution instead of channel-wise attention. It is worth noting that each angle feature convolution is followed by a $3 \times 3$ convolution and a pixel shuffle layer. The purpose of this design is twofold, one is that due to the design of the stride of the angular feature convolution, the direct output feature map decreases layer by layer, and thus an up-sampling is required to allow for the concatenation of multiple levels of angular feature convolution. And the other is that due to the fact that in the case of a large parallax, the pixels corresponding to the same object point in the space may be located in different macro-pixels, and in order to extract their parallax features it is necessary to correlate the neighboring multiple macro-pixels' features. Therefore, $3 \times 3$ convolution is used to expand the receptive field of each layer of angular feature convolution. In the original design of Wang et al. a combination of 1x1 convolution and pixel shuffle layer is used, and our experimental results show that $3 \times 3$ convolution can bring performance improvement.

\subsubsection{Spatial Feature Extraction Block (SFEB)}

The spatial feature extraction block is similar in structure to the angular feature extraction block, which contains three layers of spatial feature convolutions connected by layer-by-layer concatenation to extract the spatial information of the light field, and replaces the large convolution kernel with multiple layers of small convolution kernels connected in series to realize the receptive field enhancement. For the same consideration as the angular feature extraction, no channel-wise attention is applied. At the same time, due to the stride of 1 for spatial feature convolution, no additional auxiliary components are needed for size resizing.

We did not replace the EFE with a EPI feature extraction block for two reasons. First, we consider the EFE as a mediator of spatial and angular features, which have been adequately extracted by the corresponding extraction block. The second reason is for limiting the parameter size, as it is difficult to train a network that is too large.

\subsubsection{Deep Disentangling Block}

The DDB, as a module of the level 2, combines and packages the basic modules of the level 1 into a whole, which is the basic unit for the construction of the upper level network structure. The DDB mainly contains an AFEB, a SFEB and an EFE. The EPI feature extraction in two directions is achieved by passing the original input and the transposed input to the same EPI feature extractor, respectively. The above approach is equivalent to two polar plot feature extractors with shared parameters. The current version of the network has the same number of channels for all three of these feature extractors. A given feature in the form of an input light field macro-pixel image is denoted as $\mathcal{F}_{in} \in R^{H \times W \times C}$, where C denotes the channel dimension, and C is set to be 64 in the current version of the network. The input features are first passed through an angular feature extractor and an EPI feature extractor in parallel respectively (got outputs in both horizontal and vertical directions).The intermediate features of the 3-way outputs, denoted as $\mathcal{F}^{inter}_{ang}$, $\mathcal{F}^{inter}_{epi-h}$ and $\mathcal{F}^{inter}_{epi-v}$ are concatenated in the channel dimension and fused into a singular number of channels by a $1 \times 1$ convolution, the output is denoted as $\mathcal{F}^{inter}_{fusion}$. Afterwards, the intermediate feature $\mathcal{F}^{inter}_{fusion}$ is passed through a SFEB. The final output of the DDB is defined as the residual sum of the output of the SFEB and the $\mathcal{F}^{inter}_{fusion}$. 

\subsubsection{Deep Disentangling Block Group and Angular Up-Sampling Module}

The depth decoupled module group is the third layer of modules, which consists of a certain number of depth decoupled modules organized according to the structure of channel-by-channel splicing introduced in Section 3.1.2.

In order to save memory while increasing the depth of the network, we adopt a feature extraction-up-sampling network architecture. The angular up-sampling module is located in the last part of the network and serves to upsample the intermediate features output from the front module in the angular dimension to obtain the light field at the target angular resolution. To solve the problem of common non-integer factors (e.g., $2 \times 2 \to 7 \times 7$ ) in angular super-resolution that are not suitable for direct interpolation, we adopt the same angular up-sampling module as DistgASR. Specifically, the method employs a downsampling-upsampling strategy, where the first two dimensions of the feature $\mathcal{F} \in R^{AH \times AW \times C}$ are first reduced using an AFE to obtain $\mathcal{F}_{down} \in R^{H \times W \times C}$. Then a 1x1 convolution is used to up-sample the channel dimension of $\mathcal{F}_{down} \in R^{H \times W \times C}$ according to the input angular resolution and angular super-resolution coefficients to obtain $\mathcal{F}_{up} \in R^{H \times W \times \alpha^2 A^2 C}$. Finally, the output $\mathcal{F}_{pixel-shuffle} \in R^{\alpha A H \times \alpha A W}$ is obtained by pixel-shuffling the channel dimension to the angular dimension.

At the end of the network, the output in the form of macro-pixel images is converted to the form of sub-apertures as the final output of the network.

\section{Experiments}

\subsection{Dataset and Implement Detail}

We validated the performance of the proposed network on both synthetic and real-world scenes, respectively. The synthetic scenes are from the 4-D Light Field Benchmark\cite{hci_new, hci_old}, and the real-world scenes are from Kalantari \cite{kalantari2016learning} and STFLytro \cite{stflytro}, both captured with an Illum Lytro camera. The training and test sets are divided with reference to \cite{lfasr_fs_gaf}, specifically, there are 20 HCI new synthetic scenes and 100 real-world scenes for training, while the test set contains 4 HCI new synthetic scenes, 5 HCI old synthetic scenes, 30 real-world scenes from the 30scenes dataset, 25 scenes from $\mathit{Occlusions}$ category and 15 scenes from the Reflective category in STFlytro real-world scenes dataset. As in many previous works, we also designed the experiment as $2 \times 2 \to 7 \times 7$ angular domain super-resolution. Specifically, the high angular resolution ground truth data is a $7 \times 7$ light field obtained by cropping from the center of the angular domain in the raw data, and the input data is a $2 \times 2$ low angular resolution light field obtained by sparse sampling from the four corners of the ground truth. In the training phase, we cropped the light field data in the spatial dimension into patches of $64 \times64$ size, obtaining a total of about $7.3 \times 10^4$ training samples. Similar to the single-image super-resolution, we used random horizontal/vertical flipping and rotation for data argumentation. 

The network was trained using the L1 loss function as supervision and optimized using the Adam optimizer with parameters set to $\beta_1=0.9$, $\beta_2=0.999$. We set the batch size to 8 and the initial value of the learning rate to 2e-4, with half of it decreasing every 15 rounds, for a total of 75 rounds of training. All experiments were done on a computer with an Intel Core i7 cpu and an Nvidia RTX 3090 gpu.

We use PSNR and SSIM as quantization metrics for our experiments. Similar to other work on angular super-resolution, we transformed the images from RGB to YCbCr space and performed angular super-resolution and computed quantitative metrics for the Y channel. The metrics for each scene were computed from the average of the metrics for each new viewpoint (i.e., $7 \times 7 - 2 \times 2$ for a total of 45), and the metrics for each dataset were computed from the average of the metrics for each scene.

\subsection {Ablation Study}

In this section we will verify the effectiveness of each module or design separately by quantitative metrics of ablation experiments, including:
\begin{itemize}
    \item[(1)] Swin Transformer-like stage ratio
    \item[(2)] Channel-wise attention
    \item[(3)] $3 \times 3$ convolution in AFE
    \item[(4)] Layer-by-layer concatenation
\end{itemize}

\begin{table*}[!t]\small
\centering
\caption{Quantitative results of ablation experiments. The best results are highlighted in bold.}\label{table:1}
\resizebox{\textwidth}{10mm}{
\begin{tabular}{@{}l|llll|l|ll|l@{}}
\toprule
    NO. & Swin-T-like stage ratio & Channel-wise Attn& $3 \times 3$ Conv & Layer-by-layer Concat & Param. & \makecell[c]{HCInew} & \makecell[c]{HICold} & \makecell[c]{Avg}\\
\midrule
    \makecell[c]{1} & \makecell[c]{$\checkmark$} & \makecell[c]{$\checkmark$} & & \makecell[c]{$\checkmark$} & 30.03M & 35.11/0.971 & 42.07/0.974 & 38.59/0.973 \\
    \makecell[c]{2} & \makecell[c]{$\checkmark$} & & \makecell[c]{$\checkmark$} & \makecell[c]{$\checkmark$} & 31.90M & 35.12/0.969 & 41.79/0.971 & 38.46/0.970 \\
    \makecell[c]{3} & & \makecell[c]{$\checkmark$} & \makecell[c]{$\checkmark$} & \makecell[c]{$\checkmark$} & 32.09M & 35.08/0.967 & 41.87/0.970 & 38.48/0.969 \\
    \makecell[c]{4} & \makecell[c]{$\checkmark$} & \makecell[c]{$\checkmark$} & \makecell[c]{$\checkmark$} & & 36.05M & 34.81/0.967 & 41.73/0.969 & 38.27/0.968 \\
    Ours & \makecell[c]{$\checkmark$} & \makecell[c]{$\checkmark$} & \makecell[c]{$\checkmark$} & \makecell[c]{$\checkmark$} & 32.09M & \textbf{35.47/0.972} & \textbf{42.19/0.976} & \textbf{38.83/0.974} \\
\bottomrule
\end{tabular}}
\end{table*}

It is worth mentioning that a large number of practices have shown that the number of parameters is significantly correlated with the network performance, so the network depth is appropriately adjusted to ensure that the ablation study network after removing the network structure with a large number of parameters (e.g., $3 \times 3$ convolution in the AFE) still has a parameter number close to that of the proposed network. Similarly, for structures that reduce the number of network parameters (e.g., layer-by-layer concatenation), we appropriately reduce the depth of the ablation study network to keep the parameter scales roughly constant. The quantitative results of the ablation study is listed in Table \ref{table:1}

\subsubsection{The effectiveness of Swin Transformer-like stage ratio}

The proposed network contains 4 DDBGs, each containing several DDBs, and this ablation experiment was designed to verify the validity of the stage ratio of the number of DDBs. The ablation study for this design changes the stage ratio of 2 2 6 2 to an averaged 3 3 3 3. The ablation study network compares the effect of these two stage ratios by keeping the total number of 12 DDBGs constant, i.e., the total number of parameters in the network is constant. The results show that compared to the averaged stage ratio, a Swin Transformer-like 1:1:3:1 stage ratio is able to improve the performance in the light-field angle super-resolution task.

\subsubsection{The effectiveness of Channel-wise Attention}

The channel-wise attention module is suitable for high-level features and is often used for weighted channel compression of feature tensors with large channel dimensions. In the proposed network, the channel-wise attention module is placed as a pluggable module at the end of the level 3 and 4 modules, which serves for better channel compression. In the ablation experiments, we replace the channel-wise attention module with a $1 \times 1$ convolutional layer. We did not adjust the network depth since the channel-wise attention module has little effect on the number of parameters. The experimental results show that the channel-wise attention module is able to improve the angular super-resolution performance compared to the $1 \times 1$ convolution on larger channel counts generated by layer-by-layer concatenation.

\subsubsection{The effectiveness of 3*3 Convolution in AFE}

The $3 \times 3$ convolution in the AFE is placed after the angular feature extraction convolutional layer to recover the spatial size of the features and to fuse the angular information from different macro-pixels. This design significantly affects the number of parameters in the network, so after replacing this structure with a $1 \times 1$ convolutional layer, we increase the depth of the network appropriately to ensure that the number of parameters are roughly the same. The experimental results indicate that thanks to the larger receptive field of the $3 \times 3$ convolution, the network is able to learn the angular information between neighboring macro-pixels, which helps to improve the angular super-resolution performance of the network.

\subsubsection{The effectiveness of Layer-by-layer concatenation}

Using layer-by-layer splicing instead of the dense structure can drastically reduce the number of network parameters and make the network easier to train. In the ablation experiments, we replace all layer-by-layer splicing with the dense structure, which leads to a significant increase in the number of network parameters. The experimental results show that layer-by-layer splicing can improve the angular super-resolution performance of the network.

\subsection{Comparison with State-of-the-Art Methods}

In this section we perform a comparison experiment between the proposed method and seven sota methods in the field of light field angle super-resolution \cite{lfepicnn, sheared_epi, kalantari2016learning, yeung2018fast, lfasr, lfasr_fs_gaf, distg_asr}. We compared the average quantization results of the dataset for all the methods and compared the visualization with three of them \cite{lfasr, lfasr_fs_gaf, distg_asr}, which have the highest quantization results. The network models of the above methods were retrained on the same dataset as ours.

\begin{table*}[!t]\small
\centering
\caption{Quantitative results (PSNR/SSIM) of comparison experiments with multiple SOTA methods under $2 \times 2 \to 7 \times 7$ angular super-resolution tasks} \label{table:2}
\begin{tabular}{@{}l|lllll@{}}
\toprule
\makecell[c]{\textbf{Method}} & \makecell[c]{\textbf{\textit{HIC new}}} & \makecell[c]{\textbf{\textit{HIC old}}} & \makecell[c]{\textbf{\textit{30scenes}}} & \makecell[c]{\textbf{\textit{Occlusions}}} & \makecell[c]{\textbf{\textit{Reflective}}} \\
\midrule
\makecell[c]{\textit{LFEPICNN}\cite{lfepicnn}} & 26.64/0.744 & 31.43/0.850 & 33.66/0.918 & 32.72/0.924 & 34.76/0.930 \\
\makecell[c]{\textit{ShearedEPI}\cite{sheared_epi}} & 31.84/0.898 & 37.61/0.942 & 39.17/0.975 & 34.41/0.955 & 36.38/0.944 \\
\makecell[c]{\textit{Kalantari}\cite{kalantari2016learning}} & 32.85/0.909 & 38.58/0.944 & 41.40/0.982 & 37.25/0.972 & 38.09/0.953 \\
\makecell[c]{\textit{Yeung}\cite{yeung2018fast}} & 32.30/0.900 & 39.69/0.941 & 42.77/0.986 & 38.88/0.980 & 38.33/0.960 \\
\makecell[c]{\textit{LFASR-geo}\cite{lfasr}} & 34.60/0.937 & 40.83/0.960 & 42.44/0.985 & 38.51/0.979 & 38.48/0.956 \\
\makecell[c]{\textit{LFASR-FS-GAF}\cite{lfasr_fs_gaf}} & \textbf{37.13}/0.966 & 41.80/0.974 & 42.75/0.986 & 38.36/0.978 & 38.18/0.957 \\
\makecell[c]{\textit{DistgASR}\cite{distg_asr}} & 34.70/0.974 & 41.85/0.971 & 43.61/0.995 & 39.43/0.991 & 39.08/0.978 \\
\makecell[c]{\textit{Ours}} & 35.47/\textbf{0.974} & \textbf{42.18}/\textbf{0.976} & \textbf{44.14}/\textbf{0.996} & \textbf{40.32}/\textbf{0.993} & \textbf{39.73}/\textbf{0.981} \\

\bottomrule
\end{tabular}
\end{table*}

\subsubsection{Quantitative comparison}

The quantization experiments were all performed under the $2 \times 2 \to 7 \times 7$ angular super-resolution task and the results are shown in Table \ref{table:2}.

Of the seven methods involved in the comparison, the LFEPICNN \cite{lfepicnn} and ShearedEPI \cite{sheared_epi} methods are based on EPI, and their networks learn the correlation between the slopes of the straight lines in the polar plots and the depth of the scene in order to model the parallax structure of the light field. However, the EPI contain only one angular dimension and one spatial dimension of the 4D light field, so these two methods have difficulty in capturing sufficient and comprehensive spatial and angular information, and thus are the worst performers among all the methods involved in the comparison. Kalantari \cite{kalantari2016learning}, LFASR \cite{lfasr} and LFASR-FS-GAF \cite{lfasr_fs_gaf} explicitly estimate the scene parallax through a network in the first stage, and in the second stage, new dense views are obtained by warping and refining the input views based on the scene parallax with the stage 2 network. Instead of dividing the angular domain super-resolution into two stages of extracting parallax features and warping the input views to new dense views, Yeung \cite{yeung2018fast}, DistgASR \cite{distg_asr} and our proposed network use a deeper one-stage network to directly extract the parallax features and achieve the angular domain super-resolution by up-sampling at the end of the network. Quantitative results show that the SSIM metric of our proposed method is optimal on all five datasets and the PSNR metric is optimal on four datasets.

\subsubsection{Visual comparison}

\begin{figure*}
  \begin{center}
  \includegraphics[width=6.0in]{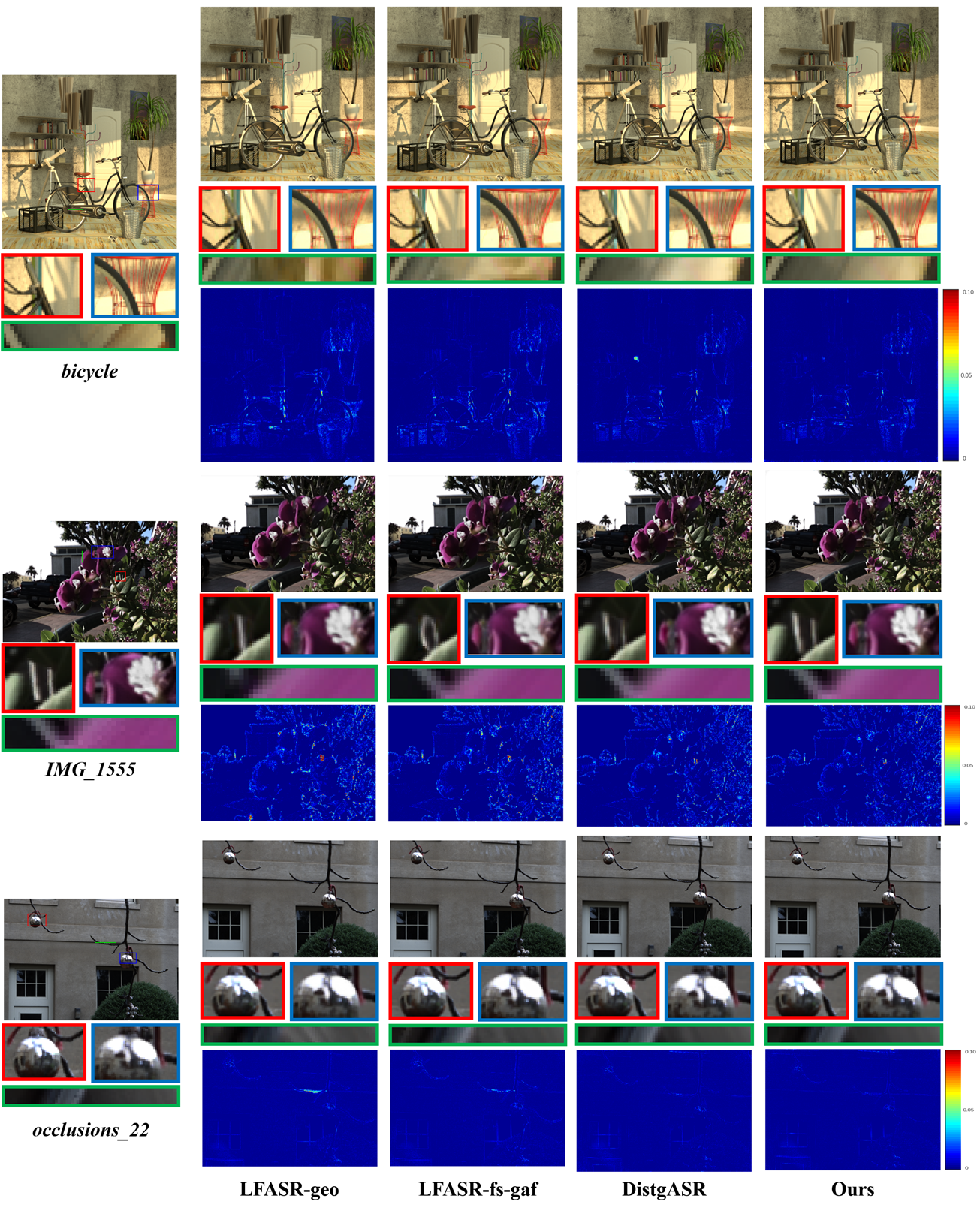}\\
 \caption{Visual comparison with LFASR-geo\cite{lfasr}, LFASR-fs-gaf\cite{lfasr_fs_gaf}, and DistgASR\cite{distg_asr} methods on the \textit{bicycle}, \textit{IMG\_1555}, and \textit{occlusions\_22} scenes. We chose the center subaperture image (CSAI) as the comparison standard, and the ground-truth image is on the left in the comparison of each scene. The upper row is the CSAI obtained by different methods. The bottom row is the error map obtained by subtracting the ground-truth image from the CSAI generated by different methods. In the middle are the 2 local regions boxed in red and blue, and the EPI corresponding to the green line.}\label{fig:visual_comp}
  \end{center}
\end{figure*}

We selected the three best performing methods, i.e., LFASR-geo \cite{lfasr}, LFASR-FS-GAF \cite{lfasr_fs_gaf}, and DistgASR \cite{distg_asr}, from the seven methods that participated in the comparison of the quantitative results for the qualitative comparison of visual effects. The visual effect comparison is shown in Fig. n. Note that due to the way the LFASR-geo \cite{lfasr} and LFASR-FS-GAF \cite{lfasr_fs_gaf} codes are implemented, there is a 22-pixel wide crop on the edges of the output results, which is therefore not completely consistent with the ground truth. The DistgASR \cite{distg_asr} and our proposed method do not have this problem. The results of visual comparision is shown in Figure \ref{fig:visual_comp}.

The visual effect comparison shows that our proposed method is closest to the ground truth image and is able to preserve the complex details in the original image. The local zoomed-in image shows that our method performs well in challenging regions such as complex textures, edge occlusions and non-Lambertian surfaces, while the EPI results show that our method is able to preserve better light field parallax structure and angular consistency.

\section {Extended applications of depth estimation enhancement}

Accurate scene depth estimation using a limited, sparsely sampled input viewpoint is an important application of light field angle super-resolution. Existing optical field depth estimation methods, whether matching-based, EPI-based or clue-based, learning-based or traditional methods \cite{sepi, wang2016depth, oavc, jin2022occlusion}, all of these methods show that the denser the input views are, the more accurate depth estimation results are obtained. Depth estimation enhancement, as an important application of light-field angle super-resolution, uses angular super-resolution methods to obtain dense views before depth estimation is performed. However, light field angular super-resolution and light field depth estimation are not seamlessly connected. The model design of light field angle super-resolution is often designed based on $2 \times 2 \to 7 \times 7$ scenarios, because the performance of angular super-resolution networks is more prominent in such extreme cases. However, in practical application scenarios of depth estimation, the input viewpoint is hardly sparse to only four corners, but a state between sparse and dense, e.g., $3 \times 3$, $5 \times 5$, etc. In this case, the angular super-resolution methods based on deep learning models often face a conflict between the huge memory usage and the limited hardware resources. Taking the $5 \times 5$ input perspective as an example, at this point, if the original network structure is still remained, the memory usage will be over 6 times of the $2 \times 2$ scene. This is unacceptable for most scenarios that do not have high memory hardware or deep learning clusters. Therefore, we need a method that can truly solve the above problem, so that angular super-resolution can be truly applied to depth estimation enhancement.

\subsection{The Block Traversal ASR Strategy}

\begin{figure}
  \begin{center}
  \includegraphics[width=3.0in]{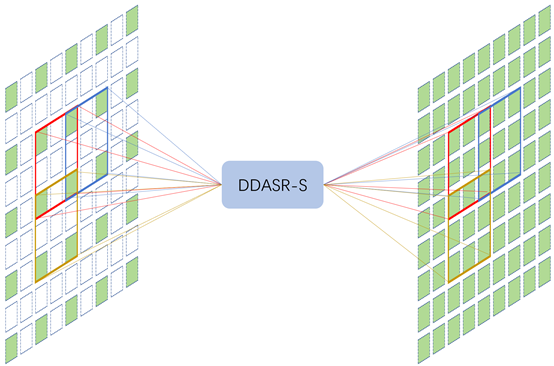}\\
 \caption{Schematic diagram of the block traversal ASR strategy for a $5 \times 5 \to 9 \times 9$ task}\label{fig:BTAS}
  \end{center}
\end{figure}

To address the above problems, we propose the Block Traversal Angular Super Resolution Strategy (BTAS). This approach can minimize the memory usage while retaining a certain network depth, so as to achieve better performance than directly reducing the network depth. Taking the $5 \times 5 \to 9 \times 9$ task as an example, as shown in Figure \ref{fig:BTAS}, we design a $2 \times 2 \to 3 \times 3$ network, and we call this network with reduced number of blocks relative to the original DDASR network DDASR-S. Because the difficulty of $2 \times 2 \to 3 \times 3$ task is lower compared to the $2 \times 2 \to 7 \times 7$ task, the size of the network it needs can be appropriately reduced. For ease of presentation and understanding, we refer to a network like DDASR-S as a Local View Network (LVN), while a network that accepts all input views simultaneously is called a Global View Network (GVN).

As shown in Figure \ref{fig:BTAS}, we use DDASR-S to traverse all blocks starting from the top left corner with a stride of 2. This stride ensures that there is an overlap with the left/up block in each slide. We divide the global parallax under the BTAS into two parts, i.e., intra-block parallax and inter-block parallax. The GVN can learn global parallax relationships from each backpropagation, while the LVN learns only the intra-block parallax relationships in the currently traversed block from each backpropagation. So an overlap is retained between each traversed block to ensure that LVNs such as DDASR-S learn the correct inter-block parallax relationships. Thus, the inter-block parallax accuracy is ensured by both the intra-block parallax of the overlapping portion and that in non-overlapping portion learned by LVN. In the case of the 2x2->3x3 task, for example, the stride of each move is 3-1=2, with 1 being the portion set aside for overlapping portion. In this way, by traversing each block with overlap, the accuracy of the inter-block parallax relationship can be propagated to all the blocks. The above process can be considered as an equivalent simulation of the GVN using the LVN.

\subsection {Experiments}

In this section, we design sufficient comparative experiments to validate the effectiveness of the BTAS using the image quality (PSNR) of the dense viewpoints obtained by the proposed method and the accuracy of depth estimation (BP1, BP7 and MSE) as a measure. We designed the experiments as using a $2 \times 2 \to 3 \times 3$ LVN (i.e., DDASR-S) in conjunction with the BTAS to achieve $5 \times 5 \to 9 \times 9$ angular super-resolution. And we performed the depth estimation enhancement based on the published method with the result of the angular super-resolution. Some other data sources are included in the comparison including $5 \times 5$ input, $9 \times 9$ angular super-resolution using GVN (memory usage is limited to the same as the LVN), and $9 \times 9$ ground truth, respectively. We also performed depth estimation based on these data sources to validate the effectiveness of the BTAS. 

\subsubsection{Network Design}

The design of DDASR-S as a LVN is broadly similar to DDASR. With the reduced task difficulty, we change the number of DDBs in the level 3 to 1, 1, 3, 1 to achieve a balance between network performance and parameter scale.

\subsubsection{Dataset and Implement Detail}

The LVN is trained using the same training set as in the previous section. We pick the central $9 \times 9$ sub-aperture images (the HCI dataset directly picks all the viewpoints, and for datasets taken by Lytro cameras such as STFLytro only the viewpoints in the center region are picked because of the poor imaging quality of the edge viewpoints), and the $3 \times 3$ local sub-aperture images with overlapping viewpoints are picked as the ground truth according to the traversal pattern of the BTAS, and the 4 corners of this block are picked as the inputs. This is done so that the trained LVN can correctly perform angular super-resolution at any localization of the global view, rather than relying on certain local position. The data preprocessing was similar to that in Section IV, with approximately $6.3 \times 10^4$ and $2.3 \times 10^5$ training samples generated on the HCI and Lytro datasets, respectively. As with DDASR, the network training was supervised with the L1 loss function and optimized by the Adam optimizer, with the Adam parameters set to $\beta_1 = 0.9$, $\beta_2 = 0.999$, the batch size increased to 12, and the initial value of the learning rate set to 2e-4, decreasing by half every 15 rounds, for a total of 75 rounds of training. All experiments were done on a computer with an Intel Core i7 processor and an Nvidia RTX 3090.
In the inference stage, according to the traversal pattern of the BTAS, the overlapping $2 \times 2$ local viewpoints are selected from the $5 \times 5$ input viewpoints, and the $3 \times 3$ viewpoints output from the local viewpoint network are used as the result for the region. For the overlapping parts, we average the overlapping parts of two neighboring blocks as the result of the overlapping parts. We use the SPO [SPO] method proposed by Zhang et al. as the final depth estimation method because of a better balance between accuracy and time-consumption.

\begin{table*}[!t]\small
% \centering
\caption{Quantitative metrics for depth estimation enhancement and angular super-resolution. } \label{table:3}
\begin{tabular}{@{}c|ccc|cccc|ccc@{}}
\toprule
& \multicolumn{3}{c}{\makecell[c]{\textit{Disparity Estimation}}} & \multicolumn{4}{c}{\textit{\makecell[c]{ASR and Disparity Estimation}}} & \multicolumn{3}{c}{\textit{\makecell[c]{Disparity Estimation}}} \\
& \multicolumn{3}{c}{\makecell[c]{$5 \times 5$ Input}} & \multicolumn{4}{c}{\makecell[c]{$9 \times 9$ (BTAS + LVN) / GVN}} & \multicolumn{3}{c}{\makecell[c]{$9 \times 9$ ground truth}} \\
& \textit{\makecell[c]{BP1}} & \textit{\makecell[c]{BP7}} & \textit{\makecell[c]{$MSE \times 100$}} & \textit{\makecell[c]{PSNR}} & \textit{\makecell[c]{BP1}} & \textit{\makecell[c]{BP7}} & \textit{\makecell[c]{$MSE \times 100$}} & \textit{\makecell[c]{BP1}} & \textit{\makecell[c]{BP7}} & \textit{\makecell[c]{$MSE \times 100$}} \\
\midrule
\textbf{boardgames} & 92.33 & 94.37 & 73.54 & 53.39/51.01 & 3.54/3.70 & 7.12/7.40 & 1.91/10.26 & 3.47 & 7.12 & 1.65 \\
\textbf{town} & 95.45 & 96.79 & 46.93 & 47.43/46.85 & 1.71/3.29 & 3.25/5.14 & 0.51/0.75 & 1.59 & 3.03 & 0.46 \\
\textbf{dishes} & 96.75 & 97.77 & 49.13 & 44.51/36.93 & 5.51/5.77 & 11.27/11.98 & 1.12/3.87 & 5.23 & 10.84 & 1.10 \\
\textbf{buddha} & 89.84 & 92.95 & 39.71 & 50.54/49.93 & 1.44/1.53 & 3.94/4.42 & 0.52/0.53 & 1.43 & 3.95 & 0.51 \\
\textbf{buddha2} & 81.75 & 85.39 & 23.95 & 44.49/44.37 & 8.37/8.89 & 13.38/14.01 & 1.17/1.55 & 7.56 & 12.48 & 1.02 \\
\textbf{monas} & 87.29 & 90.56 & 22.54 & 50.54/49.85 & 5.18/5.47 & 8.83/8.97 & 1.91/1.93 & 5.18 & 8.75 & 0.91 \\

\bottomrule
\end{tabular}
\end{table*}

\begin{figure*}
  \begin{center}
  \includegraphics[width=6.0in]{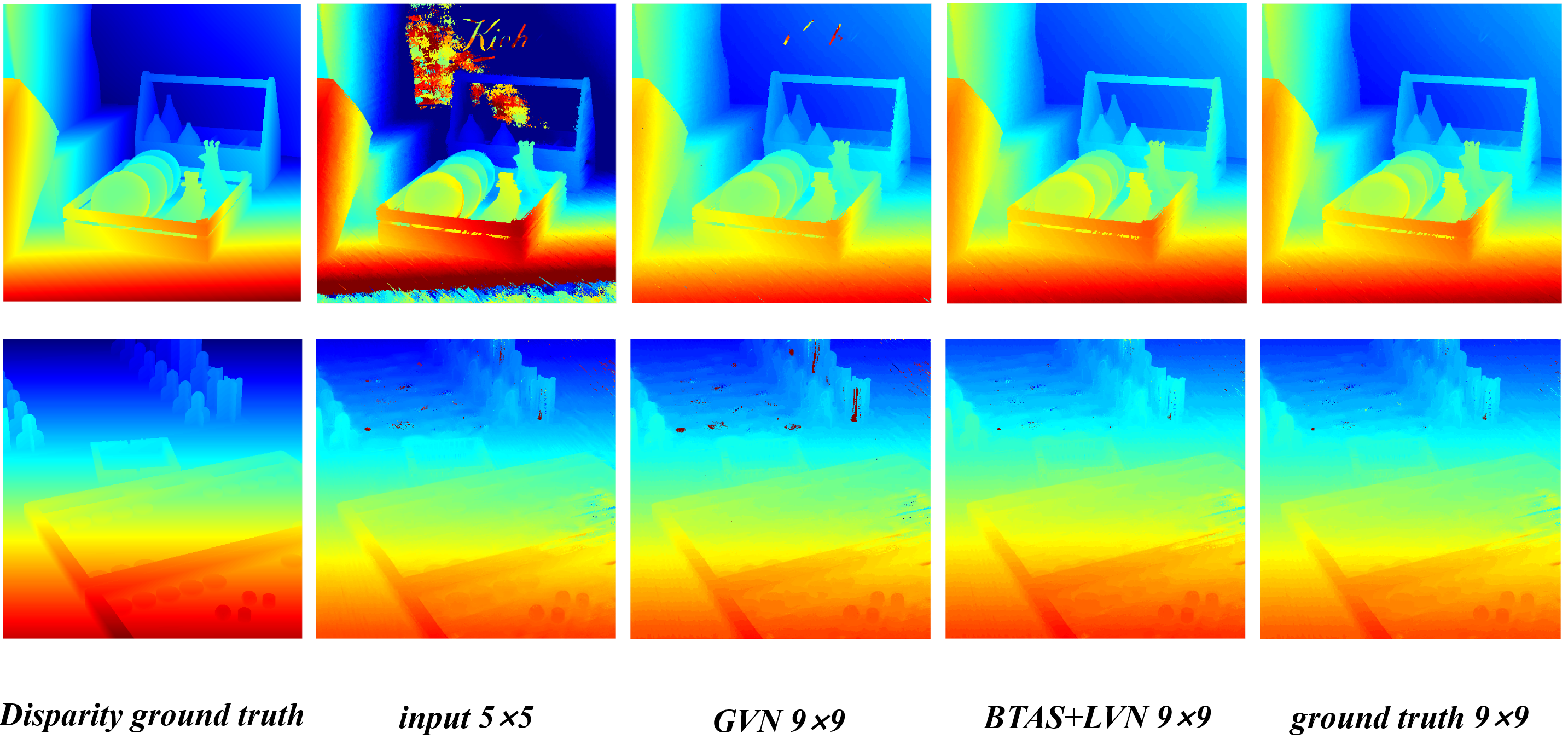}\\
 \caption{Visual comparison of depth estimation enhancement results. From left to right are the disparity map truth, the estimation results for the $5 \times 5$ input view, the depth estimation enhancement results obtained by GVN, the depth estimation enhancement results obtained by BTAS+LVN, and the depth estimation results obtained by $9 \times 9$ ground truth.}\label{fig:visual_comp_btas}
  \end{center}
\end{figure*}

\subsubsection{Experiment Results}

Since only the HCI and HCI old datasets provide the depth ground truth of the center viewpoints, our experimental validation was performed only on the two HCI datasets. The GVN also adopts a DDASR-like structure, but in order to obtain meaningful comparison results (i.e., to show the performance advantage of BTAS with a certain amount of memory usage), the GVN and LVN try to control the memory usage as similar as possible while controlling the batch size to be fixed to 8 in both cases. We limit the number of DDBs of the GVN to 2, and the final memory usage in the training stage is 16.5 Gb for the GVN and 15.7 Gb for the LVN, respectively. Table \ref{table:3} lists the quantization results of some representative scenes. BP1, BP7 and MSE were used for depth estimation enhancement, while image quality for angular super-resolution was measured by PSNR. Table \ref{table:3} shows the depth estimation metrics by column for the $5 \times 5$ input view, the image quality and depth estimation enhancement metrics for the 9x9 view generated using BTAS+LVN, the image quality and depth estimation enhancement metrics for the $9 \times 9$ view generated using GVN, and the depth estimation metrics for the $9 \times 9$ ground truth. The MSEs of the depth estimation metrics are all multiplied by 100 for ease of presentation.

Table \ref{table:3} shows that the disparity maps generated from the $5 \times 5$ input view are generally characterized by large errors, while the accuracy of the disparity maps generated from the $9 \times 9$ view obtained by both super-resolution methods is significantly improved. The results of the BTAS+LVN are higher than the GVN in all results, and slightly lower than the disparity maps generated from the $9 \times 9$ ground truth. The above quantitative results validate the feasibility and effectiveness of our proposed strategy.

We also performed a qualitative comparison of the visual effects by comparing the disparity map ground truth with the center sub-aperture disparity maps estimated from the $5 \times 5$ input view, the GVN angular super-resolution results, the BTAS+LVN angular super-resolution results, and the $9 \times 9$ sub-aperture image ground truth, respectively. The results are shown in the Figure \ref{fig:visual_comp_btas}

Qualitative comparisons of the visual effects show that the disparity maps estimated from the input viewpoints have larger errors, while the accuracy is substantially improved after angular super-resolution enhancement. Comparisons of the results obtained by GVN and BTAS+LVN show that the latter results are more accurate in some localized details, and are closer to the disparity maps obtained from the ground truth. The above qualitative comparisons are consistent with the quantitative results in Table \ref{table:3}, which together verify the effectiveness of our proposed strategy.

\section{Conclusion}
In this paper we have made several improvements to the light-field disentangling mechanism to make it better able to handle the large disparity challenges posed by sparse light fields and integrate it with advanced CNN network structures. We thus propose a deep disentangling mechanism to further improve the performance based on the effectiveness of the light-field disentangling mechanism, and design a corresponding DDASR network. We then experimentally demonstrate that the network has SOTA performance, verifying the effectiveness of the deep disentangling mechanism. Finally, we innovatively propose a block traversal angular super-resolution strategy for the problem of excessive memory usage when the input views are not that sparse, and design a lightweight DDASR-S network based on the DDASR network structure to work with the BTAS. We designed a two-stage experiment containing angular super-resolution and depth estimation to verify the effectiveness of the proposed strategy.

% or


% ============================================
%\appendices
%Appendix one text goes here %\cite{Roberg2010}.

%\section{}

%\section*{Acknowledgment}

%The authors would like to thank D. Root for the loan of the SWAP. The SWAP that can ONLY be usefull in Boulder...

\ifCLASSOPTIONcaptionsoff
  \newpage
\fi



%\begin{thebibliography}{1}


\bibliographystyle{IEEEtran}
\bibliography{IEEEabrv,Bibliography}

\vfill


\end{document}